\documentclass{article} 
\usepackage{iclr2026_conference,times}

\usepackage{amsmath,amsfonts,bm}

\def\eqref#1{equation~\ref{#1}}

\def\1{\bm{1}}

\DeclareMathAlphabet{\mathsfit}{\encodingdefault}{\sfdefault}{m}{sl}
\SetMathAlphabet{\mathsfit}{bold}{\encodingdefault}{\sfdefault}{bx}{n}

\usepackage[utf8]{inputenc} 
\usepackage[T1]{fontenc}    

\usepackage{hyperref}       
\usepackage{url}            
\usepackage{booktabs}       
\usepackage{amsfonts}       
\usepackage{graphicx}       
\usepackage{nicefrac}       
\usepackage{microtype}      
\usepackage{xcolor}         
\usepackage{makecell}
\usepackage{amsmath}

\title{Math Takes Two: A test for emergent mathematical reasoning in communication}

\author{Michael Cooper \& Sam Cooper \\
Cooper Cognitive \\
\texttt{sam(@)coopercognitive.com} \\
\texttt{michael(@)coopercognitive.com}
}

\iclrfinalcopy 
\begin{document}

\maketitle

\begin{abstract}
Although language models demonstrate remarkable proficiency on mathematical benchmarks, it remains unclear whether this reflects true mathematical reasoning or statistical pattern matching over learning formal syntax. Most existing evaluations rely on symbolic problems grounded in established mathematical conventions, limiting insight into the models' ability to construct abstract concepts from first principles. In this work, we propose Math Takes Two, a new benchmark designed to assess the emergence of mathematical reasoning through communication. Motivated by the hypothesis that mathematical cognition in humans co-evolved with the need for precise communication, our benchmark tests whether two agents, without prior mathematical knowledge, can develop a shared symbolic protocol to solve a visually grounded task where the use of a numerical system facilitates extrapolation. Unlike many current datasets, our benchmark eschews predefined mathematical language, instead requiring agents to discover latent structure and representations from scratch. Math Takes Two thus provides a novel lens through which to develop and evaluate models with emergent numerical reasoning capabilities.
\end{abstract}

\section{Introduction}

In recent years, neural network architectures, particularly large language models (LLMs), have achieved impressive performance in a wide array of tasks, including mathematical problem solving e.g. \citep{Frieder2023-oc, Lewkowycz2022-dn}. However, it is unlikely that these models enjoy the level of mathematical reasoning that we as humans do \citep{Bengio2024-kf, Rudman2025-wb}. For instance, studies highlight how LLMs fail at both compositional reasoning and mathematical tasks beyond the level of more run-of-the-mill undergraduate problems \citep{Frieder2023-oc,Ofir-Press2022-ah, Petrov2025-od}. These findings suggest that current architectures, while proficient in linguistic pattern completion, fall short of the kind of abstraction required for grounded mathematical reasoning, indicating the need for new approaches \citep{Rudman2025-wb}.

Humans, by contrast, uniquely transformed intuitive physical insights into formal, predictive mathematical models. For example, while many animals possess an intuitive sense of gravity, Newton was the first to express it through precise mathematical laws. The earliest evidence of formal arithmetic is closely tied to the emergence of record-keeping in ancient Mesopotamia \citep{Schmandt-Besserat1981-ba}. For example, Pythagorean triples appear in clay tablets used for land surveying, linking geometry to practical needs such as estimating crop yields \citep{Mansfield2021-jm}. It is plausible that even before formal notation, the drive to communicate quantities quickly and accurately, e.g., "eight attackers are approaching", created selective pressure for symbolic reasoning. Symbolic language provides dramatic advantages over repetition, e.g., saying “attacker” eight times, or relying on shared physical context. From this perspective, mathematical reasoning may have co-evolved with symbolic communication as a compact means of encoding quantitative information.

Motivated by this view and the notion that mathematics emerges through compression and abstraction \citep{Rissanen1978-ue,MacKay2003-yo,Bengio2024-kf}, we propose a new benchmark to test whether artificial agents can develop basic mathematical reasoning through communication, without access to corpora containing human language. Specifically, we challenge models to invent and extrapolate discrete symbolic representations grounded in visual stimuli, without access to predefined mathematical formalisms or human concepts of language. Our goal is being able to test the capacity of neural network architectures to rediscover counting including in $m \times n $ arrays (a potential route  to multiplication) from the bottom up, in a communicative setting.

\section{Related Work}

\textbf{Compositional language in communication: } Several studies investigate compositional generalization from a language-focused perspective. \citet{Russin2019-qb} introduce syntactic attention to improve generalization in sequence-to-sequence models on the SCAN benchmark \citep{Lake2018-nj}. Object-centric approaches also contribute to compositional generalization, especially in visual domains where disentangling viewpoint-invariant features remains a persistent challenge \citep{hinton2018matrix, locatello2020object}. More recently, benchmarks such as CoLA \citep{Ray2023-ob} evaluate alignment between compositional text and structured visual scenes. However, these studies typically assume pre-existing linguistic conventions and evaluate models within fixed symbolic systems. In contrast, our work draws inspiration from the idea that natural language did not emerge in isolation but as a tool for communication among agents. This shift in focus suggests that deeper insights into abstraction and generalization may arise from studying how structured symbolic communication can emerge interactively from grounded experience, rather than being imposed externally.

\textbf{Emergent Language in Communication Settings: } Key studies have also explored the notion that structured language emerges naturally through communication, especially in settings that mirror human language evolution. Experimental work by Verhoef and colleagues \cite{verhoef2012, verhoef2016} shows that when humans interact under memory constraints or through repeated transmission, structured signaling systems tend to emerge. These studies highlight how limitations and social coordination pressures—such as bottlenecks, alignment, and co-adaptation—can give rise to compositional structure. Similar findings have emerged from neural simulations. For example, \citet{lian2023} and \citet{zhang2024} demonstrate that linguistic features like case marking and efficient word order arise only when communication is required. Broader frameworks, including NeLLCom-X \citep{lian2024nellcomx}, and work by \citet{kouwenhoven2022}, reinforce this view by showing that language structure develops through cooperative interaction between agents grounded in shared perceptual environments. Our benchmark builds on these insights by encouraging agents to develop structured symbolic communication from scratch, under strict constraints on vocabulary and with strong generalization demands.

\textbf{Emergent language in the “Bag-Select” game: } Language emergence has often been studied in game-based scenarios, beginning with signaling games \citep{Lewis1969-cw}. Neural implementations typically involve referential setups, where a sender describes a target image among distractors \citep{Lazaridou2016-io}, or generates symbol sequences based on a single image without context  \citep{Havrylov2017-ub}. More recent work explores numerical concepts, such as \citet{Zhou2024-ay}, who introduce a visual arithmetic task but reduce it to an image-to-text mapping. Our benchmark builds most directly on the “Bag-Select” game of \citet{Guo2019-tv}, where agents communicate object quantities to guide selection. That work finds that symbolic and visual inputs better support compositional language than linguistic inputs, measured by similarity to a reference grammar. We extend this by (i) grounding all input in 2D visual object arrays, (ii) using a fixed 8-token vocabulary, and (iii) evaluating communication success directly on compositional generalization tasks. Rather than comparing to a predefined grammar, we test whether agents can construct symbolic protocols that support extrapolation to novel quantities and symbols—a hallmark of emergent reasoning  \citep{Wiedemer2023-qg}. Math Takes Two builds on this tradition by explicitly grounding symbol emergence in visual reasoning and systematically evaluating out-of-distribution generalization.

\section{Benchmarking Task}

We introduce a benchmark inspired by a hypothetical early trading scenario: two agents must communicate about quantities of goods that are not physically present, relying solely on symbolic language grounded in shared visual understanding. This setting echoes the conditions under which mathematical reasoning may have first emerged, where language enabled the abstract representation and exchange of quantities without direct sensory input.
\newline
\newline

In our benchmark, a \textit{Speaker} and a \textit{Listener}, are trained to cooperate in the following protocol:
\begin{itemize}
    \item The \textbf{Speaker} receives an input image representing a quantity of goods.
    \item The \textbf{Listener}, without access to the image, receives only a symbolic string from the Speaker and must select the correct image from a candidate set based on that string.
\end{itemize}

To limit complexity, the symbolic language is constrained to eight tokens [A, B, C, 0, 1, 2, +, *] selected to aid human performance at the task and understanding (we have already developed math from scratch) - but that should be treated uniformly by models.

This design also supports the generation of random images that can be described with strings of up to 8 characters. Crucially, the programmatic generation of these images gives researchers the option to design curricula that promote the emergence of structured communication. Example image–string pairs are shown in Figure~\ref{fig:task_overview}, details on the language used to generate images is given in Appendix \ref{app1}.

While the language in Appendix \ref{app1} represents one valid encoding of image structure, we do not enforce any fixed syntax or parsing rules. Rather, the goal is to study whether models can discover useful internal representations purely from communication and visual grounding. The benchmark consists of three phases:

\begin{figure}[t]
\centering
\includegraphics[width=1.0\linewidth]{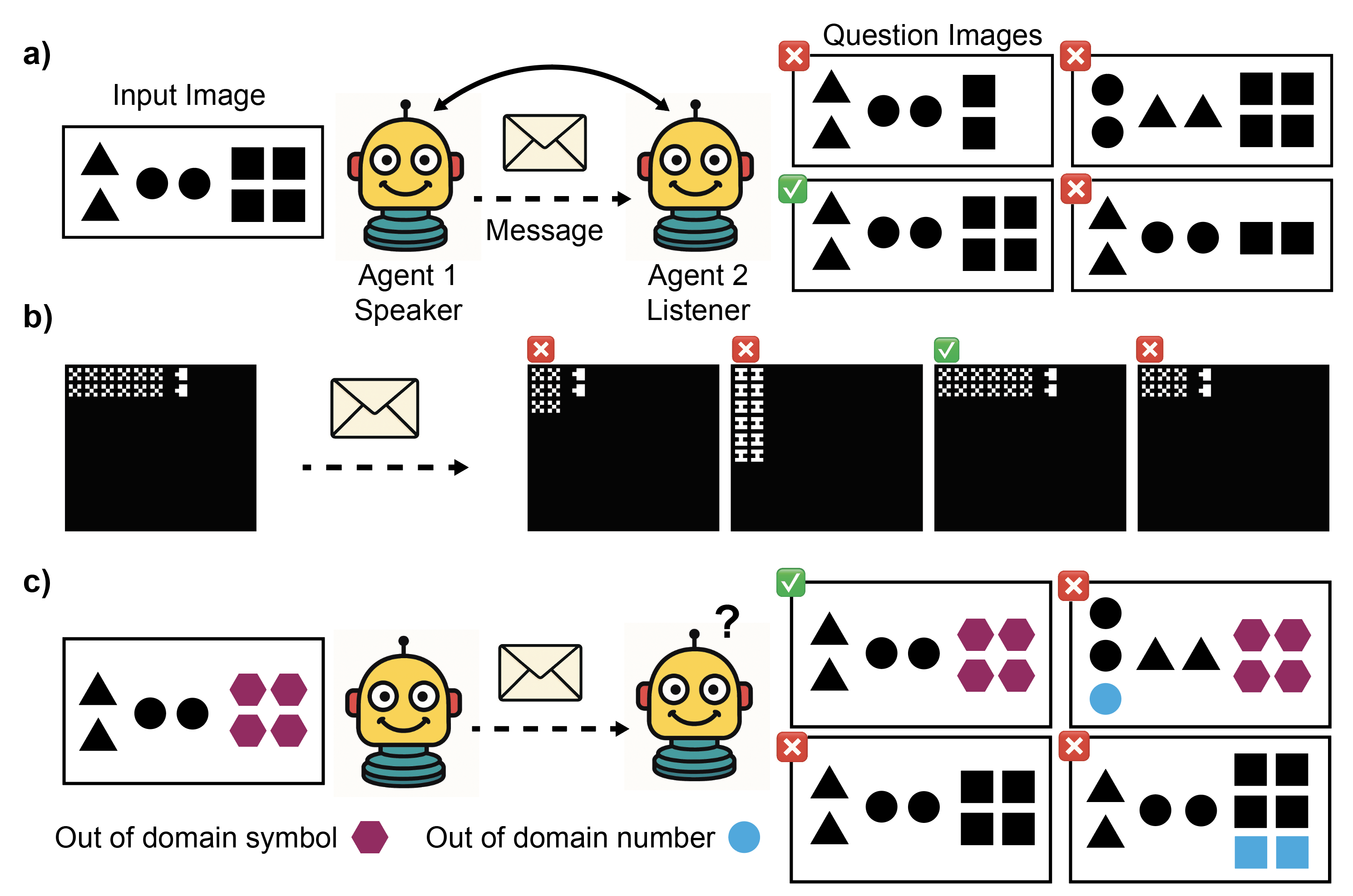}
\caption{
\textbf{Overview of the Math Takes Two benchmark.}
\textbf{(a)} The \textit{Speaker} receives an image depicting a collection of basic objects alone or in $m \times n$ arrays and communicates a symbolic string to the \textit{Listener}, who must identify the correct target image from a candidate set. In the preconditioning phase agents may interact freely and communicate bidirectionally.
\textbf{(b)} An example input image and questions set in the Math Takes Two benchmark.
\textbf{(c)} During the practice and test phases, agents encounter examples with novel object types and numbers greater than previously seen. Only one candidate image is correct. Symbolic messages are the same length and vocabulary.
}
\label{fig:task_overview}
\end{figure}

\begin{itemize}

\item \textbf{(a) Preconditioning Phase}  Agents interact freely in a shared environment, where any number of images can be generated using a limited subset of objects and numeric values. This phase is intended to facilitate unsupervised discovery of mappings between visual quantities and symbolic abstractions. Researchers are encouraged to design their own curricula during this stage.

\item \textbf{(b) Practice Phase}  Agents may proceed once sequentially through a set of 100 examples involving (i) novel object types and (ii) larger numerical values of objects than those seen in the preconditioning phase. Communication is restricted to a single symbolic string per image of at most 8 characters. The Speaker and Listener can further adapt using feedback on the response. Examples are presented in a logically ordered curriculum to facilitate deduction of new object types and numbers.

\item \textbf{(c) Test Phase}   Agents are evaluated on 100 new examples, again featuring unseen objects and quantities, but without feedback. The task requires extrapolation using one symbolic message per image, with a hard limit of 8 characters. Only one sequential forward pass is again permitted.

\end{itemize}

\textbf{Constraints} To ensure the benchmark reflects emergent vs. inbuilt reasoning, we disallow:
\begin{itemize}
    \item Models with hardcoded mathematical capabilities.
    \item Models pre-trained on mathematical datasets.
    \item Models pre-trained on natural human language corpora.
\end{itemize}

\noindent We evaluate models based on their ability to develop grounded symbolic reasoning from scratch, echoing the hypothesized origins of early mathematical abstraction in human communication. 

The Math Takes Two benchmark is available
\href{https://github.com/socooper/mathtakestwo}{\texttt{here}}.
We also provide a dataset
\href{https://huggingface.co/datasets/CooperCognitive/mathtakestwo}{\texttt{here}}.

\section{Human Performance Evaluation}

To benchmark model performance against human reasoning, we conducted a study involving 10 pairs of participants where at least one participant was required to have a technical background (undergrad engineering or sciences). Each pair was provided example images and tasked with developing a symbolic language using only the benchmark’s predefined token set: \texttt{A, B, C, 0, 1, 2, +, *}. Symbolic strings were constrained to fewer than \textit{8} characters (see Appendix \ref{app2} for further details).

Participants assumed fixed roles: the \textit{Speaker} encoded each image as a symbolic string, while the \textit{Listener} attempted to identify the corresponding image from a set of candidates using only that string. Participants were given notebooks to support the development of notation and strategy but were given no information about the image generation process or expected distribution shifts. Notebooks were submitted at the end of the exercise to evaluate the types of language systems developed by participants. The study comprised three phases:

\begin{itemize}

\item \textbf{(a) Learning Phase}  
Participants explored the environment freely, generating any number of images and developing their symbolic language collaboratively. They were also provided with a curated set of edge cases, designed to highlight the full space of possible shapes and quantities in the learning environment. No feedback or explicit instruction was given regarding underlying structure or distributional coverage.

\item \textbf{(b) Practice Phase}  
Participants were separated and given 10 structured trials. In each trial, the Speaker viewed a new image and sent a symbolic string to the Listener, who then selected an image from a candidate set. Feedback was provided after each trial, allowing refinement of their shared language. The ordering of examples was designed to support progressive generalization.

\item \textbf{(c) Test Phase}
With no further feedback, participants completed 10 new trials involving previously unseen objects and quantities. This phase assessed whether the symbolic systems developed during the practice phase could support extrapolation under novel conditions. As in the model benchmark, only a single symbolic string could be communicated per trial.
\end{itemize}

The results of the 10 pairs of participants are summarized in Table \ref{tab:results}. We include a description of the communication systems that pairs developed in Appendix \ref{app2}. Overall participants were adept at handling OOD tasks, and adapted their languages in real time to accommodate for previously unseen examples and numbers. Errors were split between specific failure modes of the developed languages, and simple errors of miscommunication including on in-domain questions.

\begin{table}
\centering
\begin{tabular}{lcccc}
\toprule
\makecell{Player} & 
\makecell{Overall \\ Accuracy} & 
\makecell{OOD Symbol \\ in Question} & 
\makecell{OOD Number \\ in Question} & 
\makecell{OOD Symbol \\ \& Number in Q.} \\
\midrule
\textit{Human Prac.} & \textbf{0.91} & \textit{0.872} & \textit{0.903} & \textit{0.933} \\

Symb AE  - F - Prac. & 0.60 & 0.532 & 0.677 & 0.625 \\
Symb AE - UF - Prac. & 0.62 & 0.617 & 0.774 & 0.750 \\
Symb Conv AE - F - Prac. & 0.77 & 0.809 & 0.613 & 0.750 \\
Symb Conv AE - UF - Prac. & 0.83 & 0.851 & 0.742 & 0.813 \\
VL Transformer - Prac. & 0.83 & 0.809 & 0.774 & 0.750 \\

\midrule

\textit{Human Test} & \textbf{0.87} & \textit{0.844} & \textit{0.731} & \textit{0.692} \\
Symb AE - F - Test & 0.60 & 0.511 & 0.538 & 0.308 \\
Symb AE - UF - Test & 0.57 & 0.444 & 0.423 & 0.231 \\
Symb Conv AE - F - Test & 0.66 & 0.578 & 0.385 & 0.615 \\
Symb Conv AE - UF - Test & 0.72 & 0.556 & 0.692 & 0.385 \\
VL Transformer - Test & 0.65 & 0.467 & 0.500 & 0.308 \\
\bottomrule
\\
\end{tabular}
\caption{
\textbf{Model and human performance in the Math Takes Two benchmark.} Accuracy is reported for both human participants and symbolic autoencoder (AE) models, under frozen (F) and unfrozen (UF) training regimes, and a vision-language transformer (VL Transformer). OOD categories contain novel symbols, numbers, or both in the question images. OOD is a new symbol or number to the specific phase (i.e., OOD in the practice phase is new vs. preconditioning, OOD in the test phase is new vs. practice.)
}
\label{tab:results}
\end{table}

\section{Baseline performance of Symbolic Autoencoders}

\begin{figure}[t]
\centering
\includegraphics[width=0.9\linewidth]{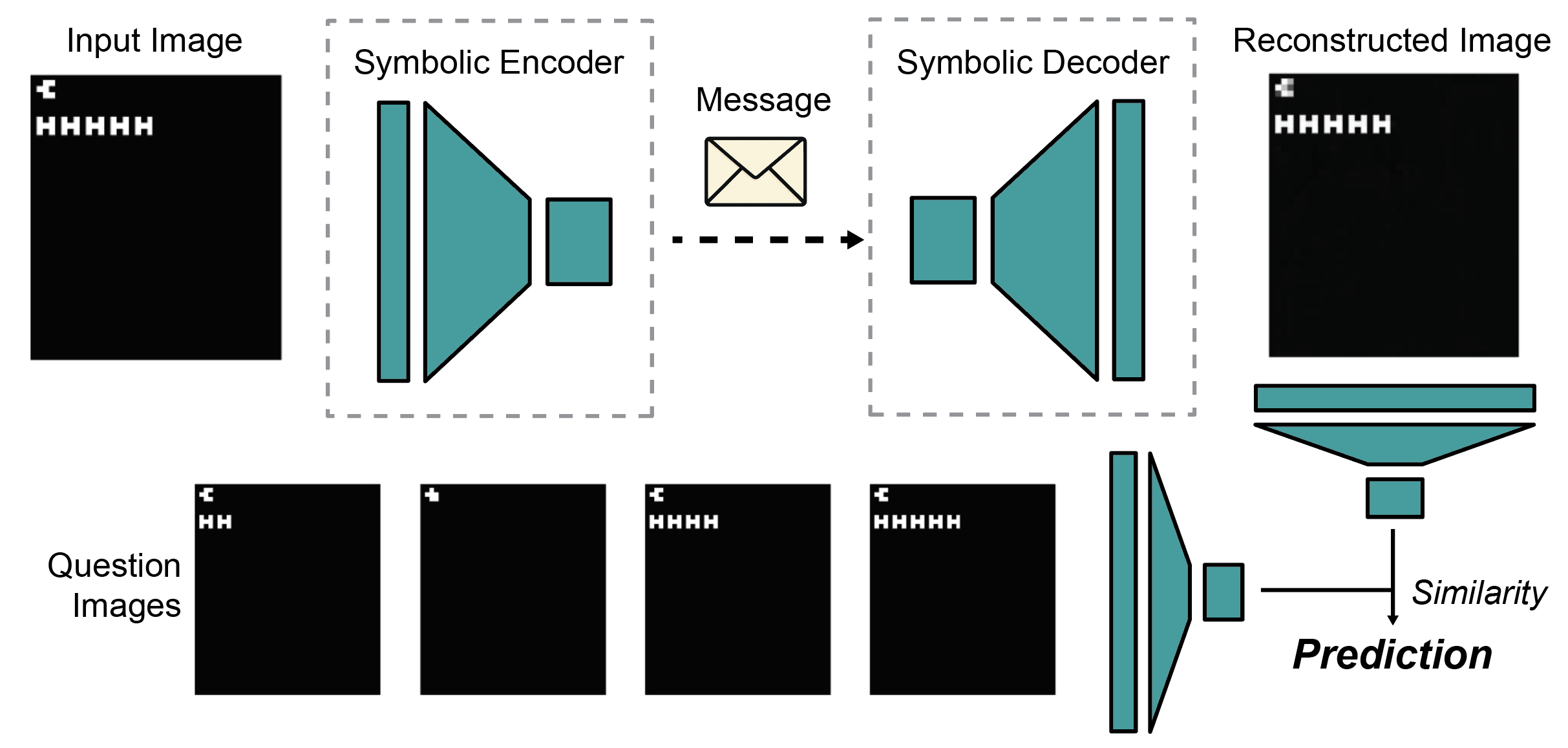}
\caption{
\textbf{Overview of the symbolic bottleneck model architecture.} The input image is first processed by a convolutional encoder that maps it to a latent feature representation. This representation is discretized into a symbolic message via a Gumbel-Softmax encoder. The message is then passed through a symbolic decoder to reconstruct the original image. For the question-answering task (bottom row), the reconstructed image is compared to a set of candidate question images using a similarity network, which produces a prediction based on the closest match.
}

\label{fig:symbolic_network}
\end{figure}

To establish a machine learning performance baseline, we evaluate two autoencoders that communicate via a \textit{symbolic bottleneck} (Figure \ref{fig:symbolic_network}), similar to the networks proposed in \citet{Guo2019-tv,Zhou2024-ay}, and \citet{Havrylov2017-ub}. Each model is trained to reconstruct an input image through this bottleneck. A secondary similarity network is then trained to predict the correct answer from a set of four options, using the reconstructed image as input (Figure \ref{fig:symbolic_network}).

The symbolic bottleneck is implemented using a Gumbel-Softmax encoder \citep{Maddison2016-xa,Jang2016-jb} that maps a flattened convolutional feature map (e.g., of shape $128 \times 5 \times 5$) into a fixed-length sequence of $L$ discrete symbols, each drawn from a categorical vocabulary of size $K$. This symbolic sequence is then decoded by a learned decoder to produce a tensor matching the original feature map dimensions. The bottleneck and decoder are optimized jointly to minimize reconstruction loss, ensuring that the symbolic representation retains essential semantic information.

The two models differ in architectural depth:

\begin{itemize}
    \item \textbf{Shallow Symbolic Autoencoder:} A single convolutional layer before symbolic encoding.
    \item \textbf{Deep Symbolic Autoencoder:} based on a fully convolutional autoencoder \cite{Masci2011-fz}, with symbolic encoding replacing the standard fully connected bottleneck after 5 warm up epochs.
\end{itemize}

Further architectural and training details are provided in Appendix \ref{app3}. We evaluate both models under two training regimes, as summarized in Table \ref{tab:results}:

\begin{itemize}
    \item \textit{Frozen:} The symbolic autoencoder is pretrained on image reconstruction using a preconditioning dataset. Its weights are then frozen, and only the similarity network is trained on the question-answering task.
    \item \textit{Unfrozen:} Both the symbolic autoencoder and the similarity network are trained end-to-end on the question-answering dataset.
\end{itemize}

Across all settings, model performance lags behind human accuracy, particularly on OOD questions. The gap is largest on examples featuring novel OOD shapes, indicating that while the symbolic bottlenecks support some generalization to new quantities and regions, they struggle to express or interpret unfamiliar visual primitives. 

Of note, we also find that while training the symbolic compression layer alongside the similarity network on the examples improves results on the practice questions, accuracy fell on the test dataset, indicating a reduction in generalization from explicitly learning to answer in-domain questions. These findings underscore the limitations of current symbolic compression schemes and motivate the development of models that can learn more generalizable communication protocols for visually grounded reasoning tasks.

\section{Limitations}

We present \textit{Math Takes Two}, a novel benchmark designed to evaluate whether two neural agents can develop emergent reasoning capabilities through communication, and whether such capabilities generalize to OOD tasks. We note three key limitations of the benchmark below.

\begin{itemize}
    \item \textbf{Shortcut-based OOD detection:} Models with strong capabilities for distinguishing in-distribution from OOD examples may achieve high performance without developing meaningful symbolic abstractions. To mitigate this, we designed specific questions that require not just identifying OOD inputs, but also selecting correct answers based on symbolic communication. Nevertheless, we cannot guarantee that language adaptation is strictly necessary for perfect accuracy.

    \item \textbf{Lack of explicit compositional evaluation:} The benchmark does not directly assess whether agents develop internal parsing mechanisms or compositional reasoning. Instead, we use performance on OOD tasks as an indirect proxy for abstraction. Naive solutions—such as hard-coded pixel-level matching across image grids—may still perform well without discovering reusable or generalizable symbolic structures. Benchmark integrity may be compromised if models exploit low-level heuristics or encode external knowledge specific to the test distribution.
    
    \item \textbf{Avoidance of pre-trained LLMs:} We also choose not to directly assess the performance of a pre-trained LLM, nor to employ an LLM teacher–student framework. This decision is intentional: the goal of the benchmark is to emphasize that mathematics was developed or discovered by humans in the absence of prior formal structures, likely emerging from evolved survival-driven cognitive biases. The core value of this benchmark, therefore, lies in understanding how analogous developmental objectives can be embedded within neural networks to enable genuine from-scratch learning.
    
\end{itemize}

\section{Future Work}

Our benchmark builds upon the “bag-selection” game proposed by \citet{Guo2019-tv}, extending it into a setting that tests both the emergence of structured symbolic communication and the capacity to generalize to examples with non-overlapping support—a gold standard for evaluating reasoning \citep{Wiedemer2023-qg}. In doing so, we aim to encourage the development of models that can describe complex visual scenes, including features related to counting, orientation, and spatial arrays.

We now know that scaling of language models to vast parameter numbers leads to dramatic improvements in performance in language and arguably reasoning \citep{vaswani2017attention}. Creating pretraining environments and scenarios that would allow large language models to develop languages in the absence of human formalisms, for example, as in \citet{lian2024nellcomx, kouwenhoven2022}, would be a key next step in trying to improve performance at this benchmark. Three behaviors observed in human players suggest further directions for improving model performance:

\begin{itemize}
    \item \textbf{Incorporating external memory mechanisms:} Participants consistently used notebooks to externalize their symbolic protocols during both the development and testing phases. This aligns with observations by \citet{Bengio2024-kf}, who argue that external memory can compensate for limited working memory and enable more advanced reasoning. These findings suggest that models equipped with constrained working memory and access to  symbolic registers or external memory modules may be better suited for this benchmark.

    \item \textbf{Modeling uncertainty and anticipating novelty:} Several participant pairs, without explicit instruction, developed fallback strategies or codes in anticipation of novel, previously unseen examples. This proactive behavior suggests that agents capable of modeling distributional uncertainty or maintaining flexible priors over possible test conditions may generalize more effectively in open-ended tasks.
    
    \item \textbf{Limitations on task breadth: } The current version of the benchmark is restricted to counting tasks over grid-based object arrays using an 8-token vocabulary. While this design isolates foundational symbolic reasoning under communication constraints, it does not yet explore more complex operations or larger vocabularies (e.g., 10 or 12 tokens). Extending the benchmark to include richer symbolic spaces and a broader range of reasoning tasks will also be an important avenue for future work.
    
\end{itemize}

\section{Conclusions}
Overall, \textit{Math Takes Two} provides a testbed for evaluating whether neural agents can move beyond statistical pattern matching to acquire structured, symbolic communication in a grounded and generalizable manner. Our framework builds on the communicative bottleneck tradition of the Bag-Select game \citep{Guo2019-tv} and related emergent language research \citep{Lazaridou2016-io}, but extends these foundations with systematic out-of-distribution (OOD) tests and visual reasoning tasks that require symbolic extrapolation. Future iterations of this framework could incorporate elementary arithmetic operations. For instance, requiring agents to match a \(bags\times objects\) input to a corresponding product in an output image. Such extensions would transform the benchmark into a complex image-to-image reasoning task, building on the symbolic counting challenges posed by \citet{Zhou2024-ay} while enforcing OOD generalization in environments devoid of predefined symbolic scaffolding.

The absence of symbolic number systems in certain human cultures provides a compelling anthropological parallel to our experimental setup. Research on isolated Amazonian communities, such as the Pirahã and Mundurukú, suggests that without a linguistic "scaffolding" for exact numbers, humans rely on an Approximate Number System (ANS) to distinguish quantities like "one," "a few," or "many" \citep{Pica2004ExactAA, Gordon2004NumericalCW}. These findings imply that while the capacity for logic may be innate, the transition from approximate perception to exact mathematical reasoning is fundamentally mediated by the development of a symbolic protocol. We hypothesize that in settings where the ecological context does not demand precise quantification, qualitative reasoning remains sufficient, and discrete symbolic concepts may never emerge. By stripping away predefined human tokens, \textit{Math Takes Two} places neural agents in a similar "pre-symbolic" state, allowing us to observe whether the functional pressure of a communicative task is sufficient to drive the shift from approximate pattern matching to exact, symbolic representation.

More broadly, our work aligns with efforts to study emergent reasoning from first principles, such as the \textit{DreamCoder} agent \citep{Ellis2020-vq, Ellis2021-xv}, which combines symbolic program induction with visual and linguistic reasoning. Inspired by this line of research, we postulate that future environments could explore even richer mathematical topographies. This might include requiring agents to negotiate protocols for communicating geometric properties like area or angle using limited auxiliary tools, such as unit-length ropes. Success in such tasks would reflect the emergence of higher-order abstractions akin to geometry and the Pythagorean theorem. Ultimately, this path leads toward machine learning systems capable of not merely executing and extending human mathematical concepts, but inventing and reasoning about novel mathematical concepts in an open-ended manner built from the ground up.

\paragraph{Ethics Statement.} Participants were recruited to take part in the study in pairs and provided informed consent for their anonymized results to be used for research purposes. No personally identifying information was collected or stored. The study involved a low-risk, educational task with no deception or intervention, and was therefore considered exempt from formal institutional ethics review.

This work investigates the emergence of symbolic reasoning in artificial agents. While advances in such capabilities may contribute to scientific understanding and the development of more general learning systems, they may also have broader societal implications, including impacts on labor and the deployment of increasingly autonomous systems. As with all research in this area, careful consideration of downstream uses and responsible deployment practices will be important as these technologies mature.

\bibliographystyle{plainnat}
\bibliography{references}

@inproceedings{Frieder2023-oc,
  title={Mathematical Capabilities of {ChatGPT}},
  author={Frieder, Simon and Pinchetti, Luca and Chevalier, Alexis and Griffiths, Ryan-Rhys and Salvatori, Tommaso and Lukasiewicz, Thomas and Petersen, Philipp Christian and Berner, Julius},
  booktitle={Advances in Neural Information Processing Systems},
  volume={36},
  pages={27699--27744},
  year={2023},
}

@ARTICLE{Lewkowycz2022-dn,
  title     = "Solving quantitative reasoning problems with language models",
  author    = "Lewkowycz, Aitor and Andreassen, Anders and Dohan, David and
               Dyer, Ethan and Michalewski, H and Ramasesh, V and Slone, Ambrose
               and Anil, Cem and Schlag, Imanol and Gutman-Solo, Theo and Wu,
               Yuhuai and Neyshabur, Behnam and Gur-Ari, Guy and Misra, Vedant",
  editor    = "Koyejo, S and Mohamed, S and Agarwal, A and Belgrave, D and Cho,
               K and Oh, A",
  journal   = "Neural Inf Process Syst",
  publisher = "Curran Associates, Inc.",
  volume    = "abs/2206.14858",
  pages     = "3843--3857",
  month     =  jun,
  year      =  2022
}

@inproceedings{Ofir-Press2022-ah,
    title = "Measuring and Narrowing the Compositionality Gap in Language Models",
    author = "Press, Ofir  and
      Zhang, Muru  and
      Min, Sewon  and
      Schmidt, Ludwig  and
      Smith, Noah A.  and
      Lewis, Mike",
    booktitle = "Proceedings of the 2023 Conference on Empirical Methods in Natural Language Processing (EMNLP)",
    year = "2023",
    pages = "5687--5711",
}

@ARTICLE{Petrov2025-od,
  title         = "Proof or bluff? Evaluating {LLMs} on 2025 {USA} math Olympiad",
  author        = "Petrov, Ivo and Dekoninck, Jasper and Baltadzhiev, Lyuben and
                   Drencheva, Maria and Minchev, Kristian and Balunović, Mislav
                   and Jovanović, Nikola and Vechev, Martin",
  journal       = "arXiv [cs.CL]",
  month         =  mar,
  year          =  2025,
  archivePrefix = "arXiv",
  primaryClass  = "cs.CL"
}

@ARTICLE{Schmandt-Besserat1981-ba,
  title     = "From tokens to tablets: A re-evaluation of the so-called
               ``numerical tablets''",
  author    = "Schmandt-Besserat, Denise",
  journal   = "Visible language",
  publisher = "search.proquest.com",
  volume    =  15,
  number    =  4,
  pages     = "321--344",
  year      =  1981
}

@ARTICLE{Mansfield2021-jm,
  title     = "Plimpton 322: A study of rectangles",
  author    = "Mansfield, Daniel F",
  journal   = "Found. Sci.",
  publisher = "Springer Science and Business Media LLC",
  volume    =  26,
  number    =  4,
  pages     = "977--1005",
  month     =  dec,
  year      =  2021,
  language  = "en"
}

@ARTICLE{Bengio2024-kf,
  title={Machine learning and information theory concepts towards an AI mathematician},
  author={Bengio, Yoshua and Malkin, Nikolay},
  journal={Bulletin of the American Mathematical Society},
  volume={61},
  number={3},
  pages={457--469},
  year={2024}
}

@BOOK{MacKay2003-yo,
  title     = "Information theory, inference and learning algorithms",
  author    = "MacKay, David J C",
  publisher = "Cambridge University Press",
  address   = "Cambridge, England",
  year      =  2003,
  language  = "en"
}

@ARTICLE{Rissanen1978-ue,
  title     = "Modeling by shortest data description",
  author    = "Rissanen, J",
  journal   = "Automatica (Oxf.)",
  publisher = "Elsevier BV",
  volume    =  14,
  number    =  5,
  pages     = "465--471",
  month     =  sep,
  year      =  1978,
  language  = "en"
}

@ARTICLE{Rudman2025-wb,
  title         = "Forgotten polygons: Multimodal Large Language Models are
                   shape-Blind",
  author        = "Rudman, William and Golovanevsky, Michal and Bar, Amir and
                   Palit, Vedant and LeCun, Yann and Eickhoff, Carsten and
                   Singh, Ritambhara",
  journal       = "arXiv [cs.CV]",
  month         =  feb,
  year          =  2025,
  archivePrefix = "arXiv",
  primaryClass  = "cs.CV"
}

@inproceedings{Guo2019-tv,
  title={The emergence of compositional languages for numeric concepts through iterated learning in neural agents},
  author={Guo, Shangmin and Ren, Yi and Havrylov, Serhii and Frank, Stella and Titov, Ivan and Smith, Kenny},
  booktitle={The Evolution of Language: Proceedings of the 13th International Conference (EvoLang13)},
  year={2020},
  publisher={Evolang}
}

@ARTICLE{Zhou2024-ay,
  title     = "Emergent communication for numerical concepts generalization",
  author    = "Zhou, Enshuai and Hao, Yifan and Zhang, Rui and Guo, Yuxuan and
               Du, Zidong and Zhang, Xishan and Song, Xinkai and Wang, Chao and
               Zhou, Xuehai and Guo, Jiaming and Yi, Qi and Peng, Shaohui and
               Huang, Di and Chen, Ruizhi and Guo, Qi and Chen, Yunji",
  journal   = "Proc. Conf. AAAI Artif. Intell.",
  publisher = "Association for the Advancement of Artificial Intelligence (AAAI)",
  volume    =  38,
  number    =  16,
  pages     = "17609--17617",
  month     =  mar,
  year      =  2024
}

@INPROCEEDINGS{Wiedemer2023-qg,
  title     = "Compositional Generalization from First Principles",
  author    = "Wiedemer, Thaddäus and Mayilvahanan, Prasanna and Bethge,
               Matthias and Brendel, Wieland",
  editor    = "Oh, A and Naumann, T and Globerson, A and Saenko, K and Hardt, M
               and Levine, S",
  booktitle = "Advances in Neural Information Processing Systems",
  publisher = "Curran Associates, Inc.",
  volume    =  36,
  pages     = "6941--6960",
  year      =  2023
}

@inproceedings{Russin2019-qb,
  title={Compositional Generalization in a Deep Seq2Seq Model by Separating Syntax and Semantics},
  author={Russin, Jake and Jo, Jason and O'Reilly, Randall C and Bengio, Yoshua},
  booktitle={Proceedings of the 2019 Workshop on Cognitive Modeling and Computational Linguistics},
  pages={52--58},
  year={2019},
  doi={10.18653/v1/W19-2907},
}

@INPROCEEDINGS{Lake2018-nj,
  title     = "Generalization without Systematicity: On the Compositional Skills
               of Sequence-to-Sequence Recurrent Networks",
  author    = "Lake, Brenden and Baroni, Marco",
  editor    = "Dy, Jennifer and Krause, Andreas",
  booktitle = "Proceedings of the 35th International Conference on Machine
               Learning",
  publisher = "PMLR",
  volume    =  80,
  pages     = "2873--2882",
  series    = "Proceedings of Machine Learning Research",
  year      =  2018
}

@ARTICLE{Ray2023-ob,
  title     = "Cola: A benchmark for compositional text-to-image retrieval",
  author    = "Ray, Arijit and Radenovic, Filip and Dubey, Abhimanyu and
               Plummer, Bryan A and Krishna, Ranjay and Saenko, Kate",
  editor    = "Oh, A and Naumann, T and Globerson, A and Saenko, K and Hardt, M
               and Levine, S",
  journal   = "Neural Inf Process Syst",
  publisher = "Curran Associates, Inc.",
  volume    =  36,
  pages     = "46433--46445",
  month     =  may,
  year      =  2023
}

@INPROCEEDINGS{hinton2018matrix,
  title     = {Matrix Capsules with EM Routing},
  author    = {Hinton, Geoffrey E. and Sabour, Sara and Frosst, Nicholas},
  booktitle = {International Conference on Learning Representations (ICLR)},
  year      = {2018}
}

@PHDTHESIS{verhoef2012,
  title  = {The Origins of Duality of Patterning in Artificial Whistled Languages},
  author = {Verhoef, Tessa},
  school = {University of Amsterdam},
  year   = {2012}
}

@ARTICLE{verhoef2016,
  title   = {Emergence of Combinatorial Structure and Economy Through Iterated Learning with Continuous Signals},
  author  = {Verhoef, Tessa and Kirby, Simon and de Boer, Bart},
  journal = {Journal of Phonetics},
  volume  = {54},
  pages   = {57--68},
  year    = {2016}
}

@ARTICLE{lian2023,
  title   = {Emergent Case Marking in Neural Agents Through Communicative Pressure},
  author  = {Lian, Ruihan and Andreas, Jacob and Zettlemoyer, Luke},
  journal = {Transactions of the Association for Computational Linguistics},
  year    = {2023}
}

@INPROCEEDINGS{zhang2024,
  title     = {Word Order Emergence in Neural Agents Requires Communication},
  author    = {Zhang, Yichi and Lian, Ruihan and Andreas, Jacob},
  booktitle = {International Conference on Learning Representations (ICLR)},
  year      = {2024}
}

@INPROCEEDINGS{lian2024nellcomx,
  title   = {NeLLCom-X: Emergent Language Learning Through Cooperative Multi-Agent Communication},
  author  = {Lian, Ruihan and Andreas, Jacob},
  booktitle = {Proceedings of the 28th Conference on Computational Natural Language Learning},
  year    = {2024}
}

@ARTICLE{Pica2004ExactAA,
  title   = {Exact and Approximate Arithmetic in an Amazonian Indigene Group},
  author  = {Pierre Pica and Cathy Lemer and V{\'e}ronique Izard and Stanislas Dehaene},
  journal = {Science},
  year    = {2004},
  volume  = {306},
  number  = {5695},
  pages   = {499--503},
  doi     = {10.1126/science.1102085},
}

@article{vaswani2017attention,
  title={Attention is all you need},
  author={Vaswani, Ashish and Shazeer, Noam and Parmar, Niki and Uszkoreit, Jakob and Jones, Llion and Gomez, Aidan N and Kaiser, {\L}ukasz and Polosukhin, Illia},
  journal={Advances in neural information processing systems},
  volume={30},
  year={2017}
}

@ARTICLE{Gordon2004NumericalCW,
  title   = {Numerical Cognition Without Words: Evidence from Amazonia},
  author  = {Peter C. Gordon},
  journal = {Science},
  year    = {2004},
  volume  = {306},
  number  = {5695},
  pages   = {496--499},
  doi     = {10.1126/science.1094492},
}

@INPROCEEDINGS{kouwenhoven2022,
  title     = {Emergence of Linguistic Structure in Cooperative Referential Games},
  author    = {Kouwenhoven, Daniel and van Steenkiste, Sjoerd and Schmidhuber, J{\"u}rgen},
  booktitle = {Advances in Neural Information Processing Systems (NeurIPS)},
  year      = {2022}
}

@INPROCEEDINGS{locatello2020object,
  title     = {Object-Centric Learning with Slot Attention},
  author    = {Locatello, Francesco and Weissenborn, Dirk and Unterthiner, Thomas and Mahendran, Aravindh and Heigold, Georg and Bachem, Olivier and Tan, David and Rae, Jack and Kohli, Pushmeet and Botvinick, Matthew},
  booktitle = {Advances in Neural Information Processing Systems (NeurIPS)},
  year      = {2020}
}

@INPROCEEDINGS{Ellis2021-xv,
  title     = "{DreamCoder}: bootstrapping inductive program synthesis with
               wake-sleep library learning",
  author    = "Ellis, Kevin and Wong, Catherine and Nye, Maxwell and
               Sablé-Meyer, Mathias and Morales, Lucas and Hewitt, Luke and
               Cary, Luc and Solar-Lezama, Armando and Tenenbaum, Joshua B",
  booktitle = "Proceedings of the 42nd ACM SIGPLAN International Conference on
               Programming Language Design and Implementation",
  publisher = "ACM",
  address   = "New York, NY, USA",
  month     =  jun,
  year      =  2021
}

@article{ellis2020-vq,
  title={Dreamcoder: growing generalizable, interpretable knowledge with wake--sleep bayesian program learning},
  author={Ellis, Kevin and Wong, Lionel and Nye, Maxwell and Sable-Meyer, Mathias and Cary, Luc and Anaya Pozo, Lore and Hewitt, Luke and Solar-Lezama, Armando and Tenenbaum, Joshua B},
  journal={Philosophical Transactions of the Royal Society A},
  volume={381},
  number={2251},
  pages={20220050},
  year={2023},
  publisher={The Royal Society}
}

@inproceedings{Lazaridou2016-io,
  title={Multi-Agent Cooperation and the Emergence of (Natural) Language},
  author={Lazaridou, Angeliki and Peysakhovich, Alexander and Baroni, Marco},
  booktitle={International Conference on Learning Representations (ICLR)},
  year={2017},
}

@ARTICLE{Havrylov2017-ub,
  title     = "Emergence of language with multi-agent games: Learning to
               communicate with sequences of symbols",
  author    = "Havrylov, Serhii and Titov, Ivan",
  journal   = "Neural Inf Process Syst",
  publisher = "proceedings.neurips.cc",
  volume    = "abs/1705.11192",
  month     =  feb,
  year      =  2017
}

@ARTICLE{Lewis1969-cw,
  title     = "Convention: A philosophical study",
  author    = "Lewis, D",
  publisher = "John Wiley \& Sons",
  year      =  1969
}

@INCOLLECTION{Masci2011-fz,
  title     = "Stacked convolutional auto-encoders for hierarchical feature
               extraction",
  author    = "Masci, Jonathan and Meier, Ueli and Cireşan, Dan and Schmidhuber,
               Jürgen",
  booktitle = "Lecture Notes in Computer Science",
  publisher = "Springer Berlin Heidelberg",
  address   = "Berlin, Heidelberg",
  pages     = "52--59",
  series    = "Lecture notes in computer science",
  year      =  2011
}

@inproceedings{Maddison2016-xa,
  title={The Concrete Distribution: A Continuous Relaxation of Discrete Random Variables},
  author={Maddison, Chris J and Mnih, Andriy and Teh, Yee Whye},
  booktitle={International Conference on Learning Representations (ICLR)},
  year={2017}
}

@inproceedings{Jang2016-jb,
  title={Categorical Reparameterization with Gumbel-Softmax},
  author={Jang, Eric and Gu, Shixiang and Poole, Ben},
  booktitle={International Conference on Learning Representations (ICLR)},
  year={2017},
  }

\appendix

\section{Specifics of the language used to develop the environment} \label{app1}

Warning: this section contains spoilers as to how to encode the images. Readers may first enjoy attempting the task as described on the github page.

\href{https://github.com/socooper/mathtakestwo/tree/main/player\_env}{https://github.com/socooper/mathtakestwo/tree/main/player\_env}

\paragraph{Symbolic Shape Language.}
We define a compact symbolic language for generating and rendering structured shape-based scenes. Each program string represents a composition of primitive shape placement commands that are parsed and visualized by a canvas-based rendering engine. Programs are syntactically valid sequences over a vocabulary of predefined shape codes and trinary-encoded layout arguments. The total alphabet consists of 8 characters, yet the maximum number of shapes encountered, and number of shape possibilities is specifically chosen to exceed 8 to challenge players to make decisions at some point as to how to use combinations of characters to represent shapes or numbers, especially out of distribution.  It is also noted that some shapes are only encoded with a single letter. These appear more frequently in the examples, though this behavior would be impossible to identify based on the number of examples given to human participants, thus perfect rediscovery of the underlying code for human players is likely impossible, and 'lossy compression' solutions are expected.

\begin{itemize}
  \item \textbf{Vocabulary:}
  \begin{itemize}
    \item Shape symbols: \texttt{A}, \texttt{B}, \texttt{C}, and all two-letter combinations from this set (e.g., \texttt{AB}, \texttt{CC}), for a total of 12 shape types.
    \item Numeric arguments: strings over the digits \texttt{0}, \texttt{1}, and \texttt{2}, interpreted as base-3 (trinary) integers (e.g., \texttt{11} = 4).
    \item Special operators: \texttt{*} horizontal or grid layout mode, \texttt{+} program concatenation (next shape command on a new region)
  \end{itemize}

  \item \textbf{Program Syntax:} Each symbolic program is a string composed of one or more layout commands concatenated with \texttt{+}. Each command is parsed into one of four rendering primitives:
  \begin{itemize}
    \item \texttt{SHAPEcc*cc}: Grid layout with \texttt{rows = parse\_trinary(cc)}, \texttt{cols = parse\_trinary(cc)}
    \item \texttt{SHAPE*cc}: Horizontal row of shapes (1 row, multiple columns)
    \item \texttt{SHAPEcc}: Vertical column of shapes (1 column, multiple rows)
    \item \texttt{SHAPE}: Single shape at origin
  \end{itemize}
  For example:
  \begin{itemize}
    \item \texttt{B12*12} → Draw shape \texttt{B} in a \(5 \times 5\) grid
    \item \texttt{BB11+AB2} → Draw \texttt{BB} in a vertical stack of 4, then draw \texttt{AB} in a vertical stack of 2 in a separate region
    \item \texttt{C} → Draw a single \texttt{C} shape
  \end{itemize}

  \item \textbf{Program Generation:} Programs are generated by a Markov model with token transitions defined over a set of states including shapes, numbers, and special operators.
  \begin{itemize}
    \item Transitions from shape tokens yield either number tokens, operator tokens (\texttt{*}, \texttt{+}), or \texttt{END}.
    \item Number tokens can be followed by layout operators or termination.
    \item Operator \texttt{*} initiates array layout and is followed by a numeric width.
    \item Operator \texttt{+} resets the parser state and begins a new shape command.
  \end{itemize}
  The generator samples valid sequences of up to 8 tokens using this transition model.

\end{itemize}

\noindent This symbolic language allows the concise generation of richly structured visual inputs, enabling studies of compositional generalization and discrete communication in neural agents.

\section{Human survey details and strategies developed } \label{app2}

Participants engaged via a set of printout pdfs provided online at 
\href{https://github.com/socooper/mathtakestwo/tree/main/player\_env}{https://github.com/socooper/mathtakestwo/tree/main/player\_env}. Participants received the \texttt{examples\_1}, \texttt{examples\_2}, and \texttt{edge\_cases} printouts. Each pair was instructed to develop a symbolic language allowing the speaker to describe the example image and the listener to select one of four answer options. During the learning phase, participants could take notes freely and were informed that in the practice phase, they would be separated. Communication would then be restricted to: (1) The speaker sending an 8-character message, (2) The listener responding with a selection, and (3) The speaker confirming correctness. In the test phase, only message passing was permitted, with results recorded afterward.

Messages could be shorter than 8 characters but could not include spaces (to reduce writing burden; padding with symbols such as \texttt{*} was possible but rarely used). Pairs typically spent 30–90 minutes developing their code, and approximately 10 minutes completing the practice and test phases. Performance data was recorded in both phases, along with observations of the communication strategies adopted. A summary of the three strategy types is provided below.

\subsection*{Player Encoding Strategies}
Participants typically adopted one of three distinct strategies when encoding images into 8-character symbolic messages drawn from an 8-symbol alphabet.

\begin{itemize}

\item \textbf{(a) Decision Tree Encoding:} Some participants employed a \textit{decision tree} strategy, where each character in the message corresponded to the answer to a categorical question about the image—for example, "What is the symbol of the most common element?" or "What is the greatest number of elements in any one direction?" Typically, two or three such questions were formulated, with each assigned to a dedicated slot in the message or separated using a spacer. All eight symbols were often used to represent predefined answers. This approach enabled participants to correctly answer many queries, including out-of-distribution (OOD) examples, due to redundancy in the encoding. For instance, even if the symbol itself was novel, the associated number might remain within the training distribution. 

\item \textbf{(b) Index-Based Encoding:} Most groups adopted an \textit{index-based} strategy, in which specific positions within the 8-character message were reserved for key attributes—typically the symbol type, row number, and column number of the first two shape groups. This lookup-style method allowed for rapid decoding and quick identification of the correct image by visual inspection. However, it proved brittle under out-of-distribution conditions, such as when the number of shapes or shape types exceeded the 8-character limit. In such cases, additional information had to be omitted or compressed, reducing the method's robustness.

\item \textbf{(c) Programmatic Encoding:} Two participant pairs employed a \textit{programmatic} strategy, aiming to reverse-engineer the underlying rules believed to govern image generation. One group adopted a base-6 naming convention for shapes and numbers, using all alphanumeric characters, with "+" to separate groups and "*" to denote operations such as shape $\times$ row $\times$ column. Another group independently rediscovered the base-3 (ternary) numerical system used in image construction and used the 8-character message to first encode shape identity, followed by ternary digits representing the total number of shapes. Although this latter approach failed to capture orientation differences, resulting in one incorrect answer during practice, it achieved high compression and led to perfect accuracy on OOD test questions, aside from a single human error on an in-distribution example.

\end{itemize}

We also recorded individual participant scores to ensure that results were reproducible across groups. These are provided on the github page in the human results table. Across the 10 player pairs, the practice mean and standard deviation were $\mu = 9.1, \sigma = 0.81$, and test mean and standard deviation were  $\mu = 8.7, \sigma = 0.56$. We considered this low standard deviation to mean ratio a sign that test results were reproducible across the set of player pairs assessed.
\section{Model Architecture and Hyperparameters} \label{app3}

\paragraph{Symbolic Autoencoder}
This model serves as our simplest baseline, encoding images directly into discrete tokens.
\begin{itemize}
  \item \textbf{Symbol Encoder:} 
    Architecture consists of \texttt{Dropout2d} $\rightarrow$ \texttt{Conv2d}(1, 64, 3) + ReLU $\rightarrow$ \texttt{AdaptiveAvgPool2d}((1, $L$)) $\rightarrow$ \texttt{Conv1d}(64, $K$, 1). It utilizes Gumbel-softmax sampling ($\tau=0.5$) to produce a symbolic matrix $[B, L, K]$, where $K=8$ (vocabulary) and $L=8$ (sequence length).
  \item \textbf{Symbol Decoder:} 
    One-hot tokens are projected to dimension $D$ via a shared linear layer. Reconstruction is performed by a two-layer MLP [\texttt{Linear}($L \cdot D$, 256) $\rightarrow$ ReLU $\rightarrow$ \texttt{Linear}(256, $C \cdot H \cdot W$)] followed by a Sigmoid activation.
\end{itemize}

\paragraph{Symbolic Convolutional Autoencoder}
A fully convolutional architecture with residual blocks used for image-to-image pre-training (50,000 examples) before symbolic bottleneck integration.
\begin{itemize}  
  \item \textbf{Backbone:} Three encoding/decoding blocks with Residual units and bilinear upsampling.
  \item \textbf{Bottleneck Replacement:} Post pre-training, the convolutional bottleneck is replaced by a 2-layer MLP encoder and 2-layer MLP decoder, interfaced via a Gumbel-softmax symbolic layer.
\end{itemize}

\paragraph{Symbolic Transformer}
This model replaces the standard bottleneck in the UNet backbone with a query-based discrete message-passing stage.

\begin{itemize}
  \item \textbf{Transformer Decoder (Image-to-Symbol):}
  \begin{itemize}
    \item \textbf{Latent Projection:} Flattened visual features are projected to $D=512$.
    \item \textbf{Query Mechanism:} $L=8$ learnable query embeddings + positional embeddings are decoded against the visual memory using a 2-layer \texttt{TransformerDecoder} ($n\_head=4$, $dropout=0.2$).
    \item \textbf{Output Heads:} $L$ position-specific heads [\texttt{Dropout} $\rightarrow$ \texttt{Linear}] generate logits for vocabulary $K=8$. Gaussian noise ($\sigma=0.1$) is added to logits during training for regularization.
  \end{itemize}
  \item \textbf{Transformer Encoder (Symbol-to-Image):}
  \begin{itemize}
    \item \textbf{Sequence Processing:} Embedded symbols ($D=128$) are processed through a 2-layer \texttt{TransformerEncoder}.
    \item \textbf{Hybrid Fusion:} A composite representation is formed by concatenating the sequence mean $[B, D]$ with the flattened sequence $[B, L \cdot D]$.
    \item \textbf{Bottleneck Reconstruction:} A final \texttt{Linear} layer projects this combined vector back to the original bottleneck shape (e.g., $32 \times 5 \times 5$).
  \end{itemize}
\end{itemize}

\paragraph{Similarity Model Architecture}
Used for the 4-way multiple-choice evaluation by computing cosine similarity in a shared latent space ($D=128$).
\begin{itemize}
  \item \textbf{Image Encoder:} \texttt{Conv2d} $\rightarrow$ ReLU $\rightarrow$ \texttt{MaxPool} $\rightarrow$ \texttt{AdaptiveAvgPool}((4, 4)) $\rightarrow$ \texttt{Linear}.
  \item \textbf{Similarity Head:} L2-normalized feature vectors compute $s_i = \cos(f_{\text{target}}, f_{q_i})$ for $i \in \{1 \dots 4\}$.
\end{itemize}

\paragraph{Hyperparameters}
\begin{itemize}
  \item \textbf{Optimization:} ADAM optimizer; Learning rate $1 \times 10^{-3}$ (pre-training) and $1 \times 10^{-4}$ (symbolic training).
  \item \textbf{Regime:} 100 epochs; Early stopping (patience 10) based on 500-example validation set.
  \item \textbf{Transformer Config:} $D=512$, $n\_heads=4$, $n\_layers=2$, $dropout=0.2$. 
  \item \textbf{Loss Functions:} MSE for reconstruction; Cross-Entropy for 4-way similarity classification.
\end{itemize}

\section{Additional Methods and LLM usage} \label{app4}

Models were all trained on AWS G4DN-2XL

vCPUs: 8, Memory: 32 GiB, GPU: NVIDIA T4

Model weights are available at: 

\href{https://huggingface.co/datasets/CooperCognitive/mathtakestwo/tree/main/checkpoints}{https://huggingface.co/datasets/CooperCognitive/mathtakestwo/tree/main/checkpoints}

An LLM was used to assist in writing the code package, at the level of code lines and blocks vs. whole  scripts and files. LLMs were used for grammar and wording suggestions only; all technical content and experimental design were authored by the researchers.
\end{document}